\documentclass[11pt,a4paper]{article}
\usepackage[hyperref]{acl2020}
\usepackage{times}
\usepackage{latexsym}
\usepackage{multirow}
\usepackage{graphicx}
\usepackage{multicol}
\usepackage{amsmath}
\usepackage{amsfonts}
\usepackage{amssymb}
\usepackage{subfig}

\usepackage{todonotes}

\usepackage{microtype}

\aclfinalcopy 

\newcommand{\eat}[1]{}
\newcommand{\E}{\mathbb{E}}
\newcommand*{\affaddr}[1]{#1} 
\newcommand*{\affmark}[1][*]{\textsuperscript{#1}}

\title{Causal Inference of Script Knowledge}

\author{%
Noah Weber \affmark[1], Rachel Rudinger \affmark[2]\affmark[3], Benjamin Van Durme \affmark[1]\\
\affaddr{\affmark[1]Johns Hopkins University}\\
\affaddr{\affmark[2]Allen Institute for Artificial Intelligence}\\
\affaddr{\affmark[3]University of Maryland, College Park}\\
}

\date{}

\begin{document}
\maketitle
\begin{abstract}
When does a sequence of events define an everyday scenario  and how can this knowledge be induced from text? Prior works in inducing such \emph{scripts} have relied on, in one form or another, measures of correlation between instances of events in a corpus.  
We argue from both a conceptual and practical sense that a purely correlation-based approach is insufficient, and instead propose an approach to script induction based on the causal effect between events, formally defined via interventions. Through both human and automatic evaluations, we show that the output of our method based on causal effects better matches the intuition of what a script represents. 
\end{abstract}

\section{Introduction}
Commonsense knowledge of everyday situations, defined in terms of prototypical sequences of events,\footnote{For simplicity we will refer to these `prototypical event sequences' as scripts throughout the paper, though it should be noted scripts as originally proposed contain further structure that is not captured here.} has long been held to play a major role in text comprehension and understanding \cite{scum1975scripts,scum77scripts,bower1979scripts,abbott1985representation}. Naturally, this has motivated a large body of work looking to learn such knowledge from text corpora through data-driven approaches. 

A minimal (and oftentimes implicit) preliminary requirement for any such approach is to provide a reasonable answer to the following: for any pair of events $e_1$ and $e_2$ what quantitative measure can be used to determine whether 
$e_2$ should follow $e_1$ in a commonsense scenario (a `script')? 

The initial work of Chambers and Jurafsky \cite{chambers:08, chambers:09} adopted point-wise mutual information (PMI) between events as an answer to the above. Later work in the same tradition employed probabilities from a language model over event sequences \cite{jans2012skip, Rudinger2015, Pichotta2016,Peng2016,weber2018hierarchical}, or other measures of event co-occurrence \cite{balasubramanian2013generating,modi-titov-2014-inducing}.

\begin{figure}[t!]
\centering
\includegraphics[scale=0.25]{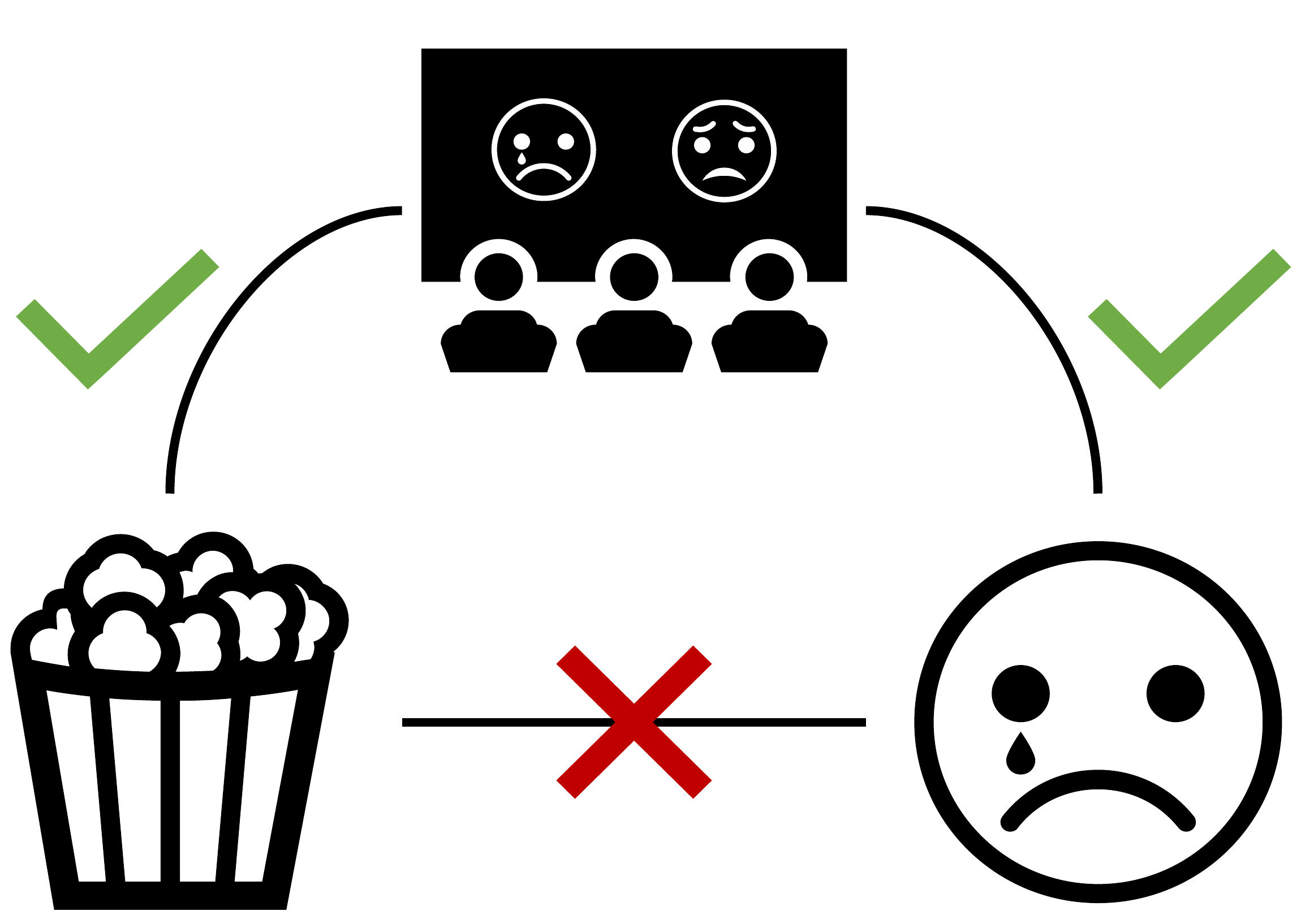}
\caption{\label{fig:cartoon}The events of \textit{Watching a sad movie}, \textit{Eating popcorn}, and \textit{Crying}, may highly co-occur in a hypothetical corpus. What distinguishes valid event pair inferences (event pairs linked in a commensense scenario; noted by checkmarks above) versus invalid inferences (noted by a `X')?} 
\end{figure}

Despite differences, these previous approaches largely follow the same underlying principle: a high enough value of the conditional probability\footnote{What is `high enough' depends on the method of course, for example, in PMI $p(e_2 | e_1)$ needs to be higher than $p(e_2)$} $p(e_2 | e_1)$ should indicate that $e_2$ should follow $e_1$ in a script.
As with any measure, introspection is required: Does a measure rooted in $p(e_2 | e_1)$ capture the notion of whether $e_2$ should follow $e_1$ in a script? We posit that it does not; observed correlations between events indicate relatedness, but relatedness is not the only factor in determining whether events form a meaningful script. 

An example given in \citet{ge2016discovering} illustrates this point: a \emph{hurricane} event may be prototypically connected with the event of \emph{donations} coming in. Likewise, the \emph{hurricane} event may also be connected to an \emph{evacuation}. But the \emph{donation} event is not connected to the \emph{evacuation} event in the same sense (and vice-versa). Nevertheless, strong statistical associations will be built between the two. Figure \ref{fig:cartoon} provides another example of this issue; clearly \emph{eating popcorn} is not linked to \emph{crying}. But if they were to co-occur together in a hypothetical corpus due to shared associations with the event of \emph{watching a sad movie}, how could a measure based on conditional probability tell the difference? In this instance, even temporal information does not provide the answer.
The problem is exacerbated when one considers that events (such as the hurricane) may not even be made fully explicit in corresponding text; they may only be strongly implied in some given context.

So what is a measure based on $p(e_2 | e_1)$ missing? In both examples, the `invalid' inferences (let's say, inferring that $e_2$=\emph{crying} is linked with $e_1$=\emph{eating popcorn}) arise from the same underlying issue; 
observing the \emph{eating popcorn} event raises the probability of \emph{crying}, \textbf{not} due to the \emph{eating popcorn} event itself, but because observing the \emph{popcorn} event implies a context of possible prior events (like \emph{watching a sad movie}), that by themselves \textit{do} raise the probability of the \emph{crying} event. To put it another way: the act of introducing a \emph{popcorn} event in an ongoing discourse would in no scenario raise the probability/degree of belief in the \emph{crying} event. Introducing the \emph{sad movie} event would. Observing the \emph{popcorn} event (or, to continue with the analogy, being told the event happened without further context) does raise this probability, but only by virtue of the shared link with the \emph{sad movie} event. Clearly, the former relationship is more inline with the type of information we wish to extract, but $p(e_2 | e_1)$ captures the later by definition.

In this paper, we argue that capturing this former relationship (does introducing $e_1$ into a discourse raise the probability of $e_2$?) is essential for any method purporting to extract this flavor of script knowledge, on both conceptual and practical grounds. 

On conceptual grounds, we posit that modeling this relationship better captures
an important property that most events linked within a classical script posses: that they be \emph{causally} linked, something underscored both in the original papers defining scripts and related works in psychology \cite{scum1975_causalchains,Black1980,trabassosperry1985}. We argue that the practical issues noted above are byproducts of ignoring this conceptual property; a mismatch between the knowledge we want to extract, and the measures we are using to extract it.

We show that this notion of `introducing $e_1$ into the discourse' 
and its resultant effects on the probability of $e_2$ can be cleanly formalized as the distribution over $e_2$  under a particular \textit{intervention}, a central object of study in the field of causal inference \cite{hernan2019causal}. We contend that measures for extracting script events from text are more aptly based on this distribution.

The exact semantics of this intervention are unambiguously specified by a 
graphical causal model of our problem ~\cite{spirtes2000causation, Pearl2000}, which we design utilizing insights from prior work in discourse processing. Under this model, we show how these intervention distributions can be defined and estimated from observational data. Using crowdsourced human evaluations and a variant of the automatic cloze evaluation, we show how this definition better captures the notion of script knowledge compared to prior standard measures, PMI and event sequence language models.

\section{Motivation} \label{sec:motivation}
Does that fact that event $e_2$ is often observed after $e_1$ in the data (i.e., $p(e_2 | e_1)$ is ``high'') mean that $e_2$ prototypically follows $e_1$, in the sense of being part of a script?  As an example of what we mean: the event of \emph{paying} is expected to follow the event of \emph{eating} while the event of \emph{running} is not.\footnote{It is commonsense that one pays for food after eating, at least in a restaurant, while running is technically possible, but would be strange.}

In this section we argue that conditional probability is not sufficient for the purpose of extracting this information from text. We argue from a conceptual standpoint that some notion of causal relevance is required. We then give examples showing the practical pitfalls that may arise from ignoring this component. Finally, we propose our intervention based definition for script events, and show how it both explicitly defines a notion of `causal relevance,' while simultaneously fixing the aforementioned practical pitfalls. 

\subsection{The Significance of Causal Relevance}
The original works defining scripts are  unequivocal about the importance of causal linkage between script events,\footnote{``\emph{...a script is not a simple list of events but rather a linked causal chain}" \cite{scum1975scripts}} and other components of the original script definition (e.g. what-ifs, preconditions, postconditions, etc.) are arguably causal in nature. Early rule-based works on inducing scripts heavily utilize intuitively causal concepts in their schema representations \cite{dejong1983acquiring, mooney}, as do related works in psychology looking at how humans store and utilize discourse information in memory \cite{Black1980, trabassosperry1985, trabassovdb85, VandenBroek1990}. 

But any measure based solely on $p(e_2 | e_1)$ is agnostic to  notions of causal relevance. Does this matter in practice? A relatively high $p(e_2 | e_1)$ indicates either: (1) a causal influence of $e_1$ on $e_2$, or (2) a common cause $e_0$ between the two, meaning the relation between $e_1$ and $e_2$ is mostly spurious. In the latter case, $e_0$ acts essentially as a \emph{confounder} between $e_1$ and $e_2$.

\newcite{ge2016discovering} acknowledges that the associations picked up by correlational measures may often be spurious (seen by the example in the intro). Their solution relies on using trends of words in a temporal stream of newswire data, and hence is fairly domain specific. In this work, we show how a more general solution may be arrived at by recognizing the problem as what it is: a confounding problem, and hence, a causal problem. 

\subsection{Defining Causal Relevance}
Early works such as \citet{scum1975scripts} are  vague with respect to the meaning of ``causally chained." Can one say that \emph{watching a movie} has causal influence on the subsequent event of \emph{eating popcorn} happening? Furthermore, can this definition be operationalized in practice? 

We argue that both of these questions may be elucidated by taking a \emph{manipulation}-based view of causation. 
Roughly speaking, this view holds that a causal relationship is one that is ``\emph{potentially exploitable for the purposes of manipulation and control}" -- \newcite{Woodward2005}. In other words, a causal relationship between \textit{A} and \textit{B} means that (in some cases) manipulating the value of \textit{A} should result in a change in the value of \textit{B}. A primary benefit of this view is that the meaning of a causal claim can be clarified by specifying what these `manipulations' are exactly. We take this approach below to clarify what exactly is meant by `causal relevance' between script events. 


Imagine an agent reading a discourse. After reading a part of the discourse, the agent has some expectations for events that might happen next. Now imagine that, before the agents reads the next passage, we surreptitiously replace it with an alternate passage in which the event $e_1$ happens. We then allow the agent to continue reading. If $e_1$ is \textit{causally relevant} to $e_2$, then this replacement should, in some contexts, raise the agents degree of belief in $e_2$ happening next 
(compared to a case where we didn't intervene to make $e_1$ happen ). 

So, for example, if we replaced a passage such that $e_1=$ \emph{watching a movie} was true, we could expect on average that the agent's degree of belief that $e_2=$ \emph{eating popcorn} happens next will be higher. In this way, we say these events are causally relevant, and are for our purposes, script events. For event pairs that are not linked in a script, the opposite is true. There exist very few contexts in which replacing the passage with the \emph{popcorn} event would raise the probability of \emph{crying}.

With this little `story,' we have clarified the conceptual notion of causal relevance in our problem, and connected it to the notion of "introducing $e_1$ into a discourse" described in the introduction. In the next section, we further formalize this story into a causal model, a necessary first step for anyone looking to compute causal effects from observed data.

\section{Method}

Here we define our causal model, show how the effects of interventions may be computed, and how these effects may be employed in extracting script-like associations between pairs of events. 

To best contrast with prior work, we use the event representation of \citet{chambers:08}.  Each event is a pair $(p, d)$, where $p$ is the event predicate (e.g. \textit{hit}), and $d$ is the dependency relation (e.g. \textit{nsubj}) between the predicate and the \emph{protagonist} entity. The protagonist is the entity that participates in every event in the considered event chain, e.g., the `Bob' in the chain `Bob sits, Bob eats, Bob pays.'

We additionally make the oft-used simplifying assumption that document order is the same as temporal order.
Future work can consider whether improvements over this assumption
can be had via models for document timeline generation such as by
\citet{govindarajan-tacl-19}

\subsection{Defining a Causal Model} \label{sec:causal_model}
A causal model defines a set of causal assumptions that are needed when computing causal quantities (such as the effect of interventions). In this paper, we use the formalism of Causal Bayesian Networks \cite{spirtes2000causation, Pearl2000}. Informally, a CBN can be thought of as a Bayesian network whose edges imply a direction of causal influence (though see both \citet{Pearl2000} and \citet{bareinboim2012local} for formal charecterizations).

Our variable of interest is the categorical variable $e_i \in E$, where $e_i$ indicates \textit{the identity of the $i^{th}$ event mentioned 
in text}, and $E$ is the set of possible atomic event types (the predicate-dependency pairs described above). It is important to note that $e_i$ does not represent a `real world' event; it is \textbf{solely a property of the text}. This interpretation of the variable $e_i$ is what is implicitly taken in prior work. 
In the context of the high level `story' given in section \ref{sec:motivation}, $e_i$ represents the identity of the event that an agent would infer upon reading the text \footnote{As we utilize automatic tools in this paper to extract the identities of events, it is important to note that there will be bias due to measurement error. Fortunately, there do exists methods in the causal inference literature that can adjust for this bias \cite{kuroki2014measurement, miao2018identifying}. \citet{wood2018challenges} derive equations in a case setting similar to ours (ie with measurement bias on the variable being intervened on). For now, we leave these efforts for future work.}

To create our causal model, we must identify the factors that play a causal role in determining the value of $e_i$. In the graph, these variables will have a directed edge incident on $e_i$. We list these variables below, along with their \textit{meaning in italics}. Variables in \textbf{bold} are those posited to have a direct causal effect on $e_i$:

\begin{figure}[t!]
\centering
\includegraphics[scale=0.5]{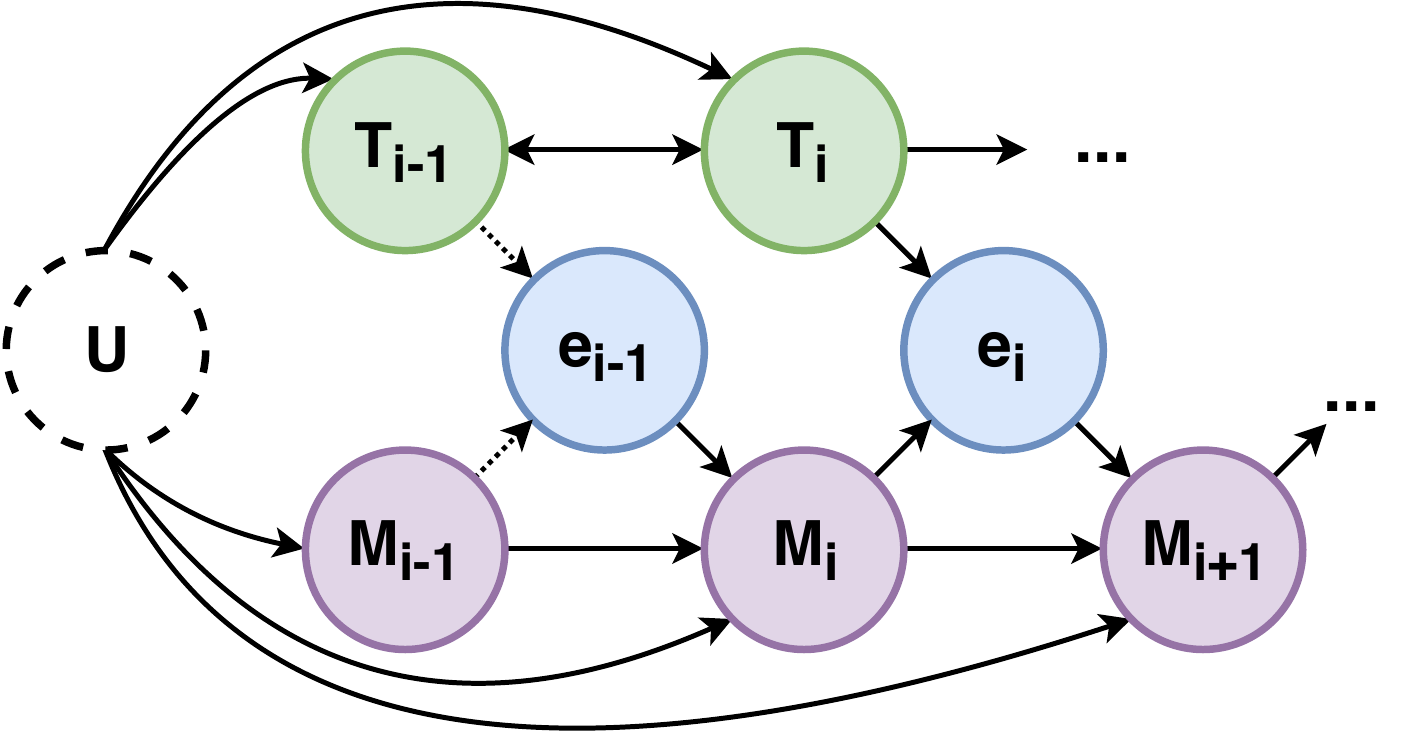}
\caption{\label{fig:causal}The diagram for our causal model up to time step $i$. Intervening on $e_{i-1}$ acts to remove the dotted edges. See \ref{sec:causal_model} for a description of the variables.} 
\end{figure}


\noindent\textbf{\emph{T}$_i$}: \textit{All the text describing the $i^{th}$ event \footnote{In this paper, we use the text output of PredPatt \cite{uds2016} as the textual representation of an event.}}. Clearly the text corresponding to $e_i$ directly affects it ($e_i$ is, after all, the identity of the event that an agent would infer after reading the text $T_i$!). However, due to the ambiguity and vagueness of text, it obviously cannot fully determine it; further context may be needed. 

The identity of $e_i$ does not only depend on $T_i$, the prior context also will also play a role in identifying $e_i$. The prior context comes in two forms: (1) the text describing prior events in the discourse, $T_0,...,T_{i-1}$ and (2) the identities of the previous events, $e_0,...,e_{i-1}$. 
It is here that we make our largest causal assumption:

\noindent\textbf{\emph{Assumption 1}}: Given the the identities of the prior events, $e_0,...,e_{i-1}$, and the current chunk of text $T_i$, the identity of $e_i$ is invariant to changes (consistent with the given values of $T_i$ and $e_0,...,e_{i-1}$) in the previous textual content, $T_0,...T_{i-1}$. In other words, \textit{the identities of the prior events capture the relevant information needed from the prior text}.  

This assumption posits that the prior text effects $e_i$ via only two causal paths; by influencing the current text that was written, $T_i$, or through its semantic content encapsulated by the identities of the prior events. Stylistic changes in $T_0,...,T_{i-1}$ that do not effect its core semantic content do not effect how we infer $e_i$, and hence, we do not include an arrow from $T_0,...,T_{i-1}$ to $e_i$.

While this assumption has intuitive appeal, it can also 
justified by prior work in 
\emph{causal network} theories of discourse processing \cite{trabassosperry1985,trabassovdb85,VandenBroek1990}. These theories hold that the causal network among events in a discourse  are a primary part of how a read discourse is represented in human memory. One could read Assumption 1 along these lines; that the prior chain of events is a sufficient representation of the prior discourse to allow reasoning about $e_i$.\footnote{Experimental results in this line of work that are especially pertinent are the priming experiments done in \citet{vanlorch1993priming} showing recognition of a particular event in a story is sped up by reminding the reader of a prior (causally related) event, as well as the experiments in \citet{trabassovdb85}, showing that surface level attributes of the text have little effect themselves on event recall}

Since we assume no direct causal influence from the prior text and $e_i$, it is clear that there must exist one between the prior events $e_0,...,e_{i-1}$ and $e_i$. For notational convenience, we represent the prior events as a single combinatorial variable $M_{i-1}$, and assume a direct arrow from $M_{i-1}$ to $e_i$. We describe this variable below:


\noindent\textbf{\emph{M}$_i$}: \textit{A variable taking a value in $2^E$ indicating all events, both described in text and left out, that happened prior to $e_i$\footnote{The set notation used here indicates that the exact ordering of the prior events in this model is arbitrary. Note that this assumption is not necessarily needed in practice (indeed, in this work, we do utilize the ordering of the prior events)}}. The prior chain of events provides the required context to, along with $T_i$, determine (up to noise) the identity of $e_i$. Note that this variable accounts for both variables described previously in the text, and those not explicitly stated in the text (out-of-text events). The value of $M_i$ is affected by $M_{i-1}$, $e_{i-1}$, and $U$, described next.

\noindent $U$: \textit{The World}. An unknowable, immeasurable variable representing the context of the world in which the text was written.

A causal diagram given in Figure \ref{fig:causal} gives a clear picture of the causal assumptions made for our problem.
Solid arrows indicate posited causal dependencies. Bidirectional arrows indicate unknown dependencies (that is, \textbf{we don't claim to know the causal dependencies between parts of the text}, any configuration is possible).

We make one other assumption for practical reasons: $M_i$ is restricted to only the previous 10 in-text events, and only contains out of text (inferred) events from step $i$.
\subsection{Identifying Intervention Distributions}
As specified by our story in section \ref{sec:motivation}, our goal is to compute the effect that intervening and setting the proceeding event $e_{i-1}$ to $k\in E$ has on the distribution over the subsequent event $e_i$. Now that we have a causal model in the form of \ref{fig:causal}, we can now meaningfully define this quantity. Using the notation of \newcite{Pearl2000}, we write this as:
\begin{align}
p(e_i | do(e_{i-1}=k))\label{eq:intervention}
\end{align}
The semantics of $do(e_{i-1}=k)$ are defined with respect to the graph, corresponding to a graph in which we have deleted the incoming arrows of $e_{i-1}$ and set it to $k$ (the dotted arrows in Figure \ref{fig:causal}). Before a causal query such as Eqn.~\ref{eq:intervention} can be estimated
we must first establish identifiability \cite{shpitser2008complete}: can the causal query be written as a function of (only) the observed data?

Eqn.~\ref{eq:intervention} is identified by noting that variables $T_{i-1}$ and $M_{i-1}$ meet the `back-door criterion' of \citet{pearl1995causal}, allowing us to write Eqn.~\ref{eq:intervention} as the following:
\begin{align}
\E_{T_{i-1}, M_{i-1}}\Big[p(e_i | e_{i-1}=k, M_{i-1}, T_{i-1})\Big]  \label{eq:est}
\end{align} 

Our next step will be estimating the above equation. If one has a estimate for the conditional $p(e_i | e_{i-1}, M_{i-1}, T_{i-1})$, then one may plug it into Eq \ref{eq:est} and use a Monte Carlo estimate to approximate the expectation (using samples of $(T,M)$ from our dataset). This leads to an simple estimator called a \textit{plugin estimator}, and is what we utilize here. 

It is important to be aware of the fact that this estimator, specifically when plugging in machine learning methods, is quite naive (eg \citet{chernozhukov2018double}), and will suffer from an asymptotic (first order) bias \footnote{See \citet{fisher2018visually} for an introduction on how this bias manifests.} which prevents one from constructing meaningful confidence intervals or performing certain hypothesis tests. That said, in practice these machine learning based plug in estimators can achieve quite reasonable performance  (see for example, the results in \citet{shalit2017estimating}), and since our current use case can in some sense be validated empirically (quite the rare occurrence), we save the utilization of more sophisticated estimators for future work.

\subsection{Estimating the Needed Conditional} \label{sec:outoftext}
Eq~\ref{eq:est} has a dependency on the conditional, $p_{e_i} = p(e_i | e_{i-1}, M_{i-1}, T_{i-1})$, which we estimate via standard machine learning techniques using a dataset of samples drawn from $p(e_i, e_{i-1}, M_{i-1}, T_{i-1})$. There are two issues to deal with here: (1) How to deal with out-of-text events in $M_{i-1}$? (2) What form will $p_{e_i}$ take?
\paragraph{Dealing with Out-of-Text Events}
Recall that $M_i$ is `bag' of all previous events, both those that occur in the text, $M_i^I$, and those that are implicit and not in the text, $M_i^O$, such that $M_i = M_i^I \cup M_i^O$. To learn a model for $p_{e_i}$ we require samples from the full joint (which includes $M_i^O)$, though we  only have access to $p(e_i, e_{i-1}, M_{i-1}^I, T_{i-1})$. If, for the samples in our current dataset, we could draw samples from $p_{M}=p(M_{i-1}^O | e_i, e_{i-1}, M_{i-1}^I, T_{i-1})$, we would result in a dataset with samples drawn from the full joint. 

In order to `draw' samples from $p_{M}$ we employ human annotation. Annotators are presented with a human readable form of $(e_i, e_{i-1}, M_{i-1}^I, T_{i-1})$\footnote{In the final annotation experiment, we found it easier for annotators to be only provided the text $T_{i-1}$, given that many events in $M_{i-1}^I$ are irrelevant.} and are asked to annotate for possible events belonging in $M_{i-1}^O$. Rather than opt for noisy annotations obtained via freeform elicitation, we instead provide users with a set of 6 candidate choices for members of $M_{i-1}^O$. The candidates are obtained from various knowledge sources: ConceptNet \cite{Speer2012}, VerbOcean\cite{verbocean}, and high PMI events from the NYT Gigacorpus \cite{graff2003english}. The top two candidates are selected from each source.

In a scheme similar to \citet{Zhang2017}, we ask users to rate candidates on an ordinal scale and consider candidates rated above a certain value to be considered within $M_{i-1}^O$. We found annotator agreement to be quite high, with a Krippendorf's $\alpha$ of 0.79. Under this scheme, we crowdsourced a dataset of 2000 fully annotated examples on the Mechanical Turk platform. An image of our annotation interface is provided in the Appendix.

\paragraph{The Conditional Model}
We opt to use neural networks to model $p_{e_i}$. In order to deal with the small amount of fully annotated data available, we employ a finetuning paradigm. We first train a model on a large dataset that does not include annotations for $M_{i-1}^O$. This model consists of a single layer, 300 dimensional GRU encoder which encodes $[M_{i-1}^I, e_{i-1}]$ into a vector $v_e \in R^d$ and a CNN-based encoder which encodes $T_{i-1}$ into a vector $v_t \in R^d$. We then model $p_{e_i}$ as 
\[
p_{e_i} \propto Av_e + Bv_t
\]
for matrices $A$ and $B$ of dimension $|E| \times d$. 
We finetune this model on the 2000 annotated examples including $M_{i-1}^O$, leading to the model:
\[
p_{e_i} \propto Av_e + Bv_t + Cv_o
\]
where $v_o$ is the average of the embeddings for the events found in $M_{i-1}^O$ and $C$ is a new parameter matrix with the same dimensions as $A$ and $B$. Everything else is defined as before. See Appendix for further training details.

\subsection{Extracting Script Knowledge}
Provided a model of the conditional $p_{e_i}$ we can estimate $p(e_i | do(e_{i-1}=k))$ by Eq \ref{eq:est}. We evaluate the expectation by Monte Carlo, taking our annotated dataset of $N=2000$ examples and computing the following average:
\begin{align}
C_k = \frac{1}{N} \sum_{j=1}^N p(e_i | e_{i-1}=k, M_j, T_j)\label{eq:avg}
\end{align}
Which gives us a vector $C_k\in R^{|E|}$ whose $l^{th}$ component, $C_{kl}$ gives $p(e_i=l | do(e_{i-1}=k))$. We compute this vector for all values of $k$ (this computation only needs to be done once).

There are several ways one could extract script-like knowledge using this information. In this paper, we define a normalized score over intervened-on events such that the script compatibility score between two concurrent events is defined as:
\begin{align}
S(e_{i-1}=k, e_i=l) = \frac{C_{kl}}{\sum^E_{j=1} C_{jl}}\label{eq:score}
\end{align}

\section{Experiments and Evaluation}
Automatic evaluation of methods that extract script-like knowledge is an open problem that we do not attempt to tackle here,\footnote{See discussions by \newcite{Rudinger2015} and \newcite{chambers2017behind}.} relying foremost on crowdsourced human evaluations to validate our method. 

However, as we aim to provide a contrast to prior script-induction approaches, we perform an experiment looking at a variant of the popular, but knowingly flawed~\cite{chambers2017behind} automatic narrative cloze evaluation, in which the cloze test set is increasingly filtered to remove instances who's answer are high frequency events.

\subsection{Dataset}
For these experiments, we use the Toronto Books corpus \cite{zhu2015aligning,NIPS2015_5950}, a collection of fiction novels spanning multiple genres.
The original corpus contains 11,040 books by unpublished authors.
We remove duplicate books from the corpus (by exact file match), leaving a total of 7,101 books.
The books are assigned randomly to train, development, and test splits in 90\%-5\%-5\% proportions.
Each book is then run through a pipeline of tokenization with CoreNLP 3.8 \cite{manning-EtAl:2014:P14-5}, parsing with CoreNLP's universal dependency parser \cite{NIVRE16.348} and coreference resolution \cite{clark-manning-2016-improving}, before feeding the results into PredPatt \cite{uds2016}. We additionally tag the events with factuality predictions from \citet{rudinger2018neural} (we only consider factual events). The end result is a large dataset of event chains centered around a single protagonist entity, similar to \cite{chambers:08}. We make this data public to facilitate further work in this area. See the Appendix for a full detailed overview of our pipeline.\footnote{Though not used in experiments here, we also annotate all event arguments with semantic proto-role properties output by \citet{rudinger-etal-2018-neural}, which will be similarly made public.}
\subsection{Baselines}
In this paper, we compare against the two dominant approaches for script induction (under a atomic event representation\footnote{There are also a related class of methods based on creating compositional event embeddings \cite{modi2016event, weber2018event}. Since the event representation used here is atomic it makes little sense to use them here.}): PMI (similar to \citet{chambers:08, chambers:09}) and LMs over event sequences \cite{Rudinger2015, Pichotta2016}. We defer definitions for these models to the cited papers, below we provide the relevant details for each baseline, with further training details provided in the Appendix.

For computing PMI we follow many of the details from \cite{jans2012skip}. Due to the nature of the evaluations, we utilize their `ordered PMI' variant. Also like \citet{jans2012skip},  we use skip-bigrams with a window of 2 to deal with count sparsity. Consistent with prior work we additionally employ the discount score of \citet{pantel2004automatically}. For the LM, we use a standard, 2 layer, GRU-based neural network language model, with 512 dimensional hidden states, trained on a log-likelihood objective.

\subsection{Eval I: Pairwise Event Associations} \label{sec:pairwise}


\begin{table}[t!]
\centering
\begin{tabular}{c|c|c} 
Method & Average Score & Average Rank (1-6) \\
 \hline
 Causal & 49.71 & 4.10  \\
 LM & 35.95 & 3.39  \\
 PMI & 34.92 & 3.02  \\

\end{tabular}
\caption{Average Annotator Scores in Pairwise annotation experiment}
\label{tab:pairwise_he}
\end{table}

Any system aimed at extracting script-like knowledge should be able to answer the following \emph{abductive} question: given an event $e_i$ happened, what previous event $e_{i-1}$ best explains why $e_i$ is true? In other words, what $e_{i-1}$, if it were true, would maximize my belief that $e_i$ was true. We evaluate each method's ability to do this via a human evaluation. 

\begin{table}[t!]
\centering
\begin{tabular}{c|c|c||c} 
 Causal  & LM  & PMI  & Target  \\
 \hline
 \textit{X tripped} & \textit{X came} & \textit{X featured} & \textit{X fell} \\
 \textit{X lit} & \textit{X sat} & \textit{X laboured} & \textit{X inhaled} \\
 \textit{X aimed} & \textit{X came} & \textit{X alarmed} & \textit{X fired} \\
 \textit{X poured } & \textit{X nodded} & \textit{X credited} & \textit{X refilled} \\
 \textit{X radioed} & \textit{X made} & \textit{X fostered} & \textit{X ordered} \\
 
\end{tabular}
\caption{Examples from each system, each of which outputs a previous event that maximizes the score/likelihood that the Targeted event follows  in text.}
\label{tab:example_out}
\end{table}

On each task,  annotators are presented with six event pairs $(e_{i-1}, e_i)$, where $e_i$ is the same for all  pairs, but $e_{i-1}$ is generated by one of the three systems. Similar to the human evaluation in \citet{Pichotta2016}, we filter out outputs in the top-20 most frequent events list for all systems. For each system, we pick the top two events that maximize $S(\cdot, e_i)$, $PMI(\cdot, e_i)$, and $p_{lm}(\cdot, e_i)$, for the Causal, PMI, and LM systems respectively, and present them in random order. For each pair, users are asked to provide a scalar annotation (from 0\%-100\%, via a slider bar) on the chance that $e_i$ is \textit{true afterwards or happened as a result of} $e_{i-1}$. The annotation scheme is modeled after the one presented in \citet{sakaguchi2018efficient}, and shown to be effective for paraphrase evaluation in \citet{hu2019parabank}. Example outputs for systems are provided for several $e_1$ choices for this task in Table \ref{tab:example_out}.


The evaluation is done for 150 randomly\footnote{Note that we do manually filter out of the initial random list events which we judge as difficult to understand} chosen instances of $e_i$, each with 6 candidate $e_{i-1}$. We have two annotators provide annotations for each task, and similar to \citet{hu2019parabank}, average these annotations together for a gold annotation.

In Table \ref{tab:pairwise_he} we provide the results of the experiment, providing both the average annotation score for the outputs of each system, as well as the average relative ranking (with a rank of 6 indicating the annotators ranked the output as the highest/best in the task, and a rank of 1 indicating the opposite). We find that annotators consistently rated the Causal system higher. The differences (in both Score and Rank) between the Causal system and the next best system are significant under a Wilcoxon signed-rank test $(p < 0.01)$.

\subsection{Eval II: Event Chain Completion}\label{sec:chain}
\begin{table}[t!]
\centering
\begin{tabular}{c|c|c} 
Method & Average Score & Average Rank (1-3) \\
 \hline
 Causal & 60.12 & 2.19  \\
 LM & 57.40 & 2.12  \\
 PMI & 44.26 & 1.68  \\

\end{tabular}
\caption{Average Annotator Scores in Chain annotation experiment}
\label{tab:chain_he}
\end{table}

Of course, while pairwise knowledge between events is a minimum prerequisite, we would also like to generalize to handle chains of events containing multiple events (in our case, essentially equivalent to the `narrative chains' studied in \citet{chambers:08}). In this section, we look at each system's ability to provide an intuitive completion to an event chain. More specifically, the model is provided with a chain of three context events, $(e_1, e_2, e_3)$, and is tasked with providing a suitable $e_4$ that might follow given the first three events.
We evaluate each method's ability to do this via a human evaluation.

Since both PMI and the Causal model \footnote{Generalizing the Causal model to multiple interventions, though out of scope here, is a clear next step for future work.} work only as pairwise models, we adopt the method of \citet{chambers:08} for chains. For both the PMI and Causal model, we pick the $e_4$ that maximizes $\frac{1}{3}\sum^3_{i=1} Score(e_i, e_4)$, where $Score$ is either $PMI$ or Eq \ref{eq:score}. The LM model chooses an $e_4$ that maximizes the joint over all events.

Our annotation task is similar to the one in \ref{sec:pairwise}, except the pairs provided consist of a context $(e_1, e_2, e_3)$ and a system generated $e_4$. Each system generates its top choice for $e_4$, giving annotators 3 pairs\footnote{We found providing six pairs per task to be overwhelming given the longer context} to annotate for each task (i.e. each context). On each task, human annotators are asked to provide a scalar annotation (from 0\%-100\%, via a slider) on the chance that $e_4$ is \textit{true afterwards or happened as a result of} the chain of context events. 
The evaluation is done for 150 tasks, with two annotators on each task. As before, we average these annotations together for a gold annotation.

In Table \ref{tab:chain_he} we provide results of the experiment. Note the the rankings are now from 1 to 3 (higher is better). We find annotators usually rated the Causal system higher, though the LM model is much closer in this case. The differences (in both Score/Rank) between the Causal and LM system outputs are not significant under a Wilcoxon signed-rank test, though the differences between the Causal and PMI system is $(p < 0.01)$. The fact that the pairwise Causal model is still able to (at minimum) match the full sequential model on a chain-wise evaluation speaks to the robustness of the event associations mined from it, and further motivates work in extending the method to the sequential case. 

\subsection{Diversity of System Outputs}
But what type of event associations are found from the Causal model? As noted both in \citet{Rudinger2015} and in \citet{chambers2017behind}, PMI based approaches can often extract intuitive event relationships, but may sometimes overweight low frequency events or suffer problems from count sparsity. LM based models, on the other hand, were noted for their preference towards  boring, uninformative, high frequency events (like 'sat' or 'came'). So where does the Causal model lay on this scale?

We study this by looking at the percentage of unique words used by each system in the previous evaluations, presented in Table \ref{tab:newwords}. Unsurprisingly, we find that PMI chooses a new word to output often (77\%-84\% of the time), while the LM model very rarely does (only 7\%-13\%). The Causal model, while not as adventurous as the PMI system, tends to produce very diverse output, generating a new output 60\%-76\% of the time. Both the PMI and Causal system produce relatively less diverse output on the chain task, which is expected due to the averaging scheme used by each to select events.

Qualitatively looking at the output, it appears that the Causal model indeed produces answers similar to the `good' outputs of PMI system, while also being more robust to noise due to sparse counts. The top two most output events of each system for both annotations are provided to illustrate this in Table \ref{tab:topwords}. See also the model outputs in Table \ref{tab:example_out}.

\begin{table}[t!]
\centering
\begin{tabular}{c|c|c} 
Method & Pairwise & Chain \\
 \hline
 \multirow{2}{*}{Causal} & \textit{X awoke} (2\%) & \textit{X collided} (4\%)  \\
 & \textit{X parried} (1\%) & \textit{pinched X} (3\%)\\
 \hline
  \multirow{2}{*}{LM} &  \textit{X came} (30\%) & \textit{X made} (23\%)  \\
 & \textit{X sat} (27\%) & \textit{X came} (15\%)  \\
 \hline
  \multirow{2}{*}{PMI} & \textit{X lurched} (1\%) & \textit{bribed X} (3\%)  \\
  & \textit{X patroled} (1\%) & \textit{X swarmed} (2\%)  \\
\end{tabular}
\caption{Two most used output events (and \% of times it is used) for each system, for each human evaluation}
\label{tab:topwords}
\end{table}

\begin{table}[t!]
\centering
\begin{tabular}{c|c|c} 
Method & Pairwise & Chain \\
 \hline
 Causal & 76.0\% & 60.1\%  \\
 LM & 7.30\% & 13.3\%  \\
 PMI & 84.0\% & 77.6\%  \\

\end{tabular}
\caption{\% of times a system outputs a new event it previously had not used before.}
\label{tab:newwords}
\end{table}

\subsection{Infrequent Narrative Cloze}

\begin{table*}[h!]
\centering
\begin{tabular}{|c|c|c|c|c|c|c|c|} 
\hline
\multirow{2}{*}{Method}& \multicolumn{7}{|c|}{Exclusion Threshold} \\
\cline{2-8}
 &  $<0$ & $<50$ & $<100$ & $<125$ & $<150$ & $<200$ & $<500$\\
 \hline
Causal & 5.60 & 7.10 & 7.00 & \textbf{7.49} & \textbf{7.20} & \textbf{8.20} & \textbf{9.10}\\ 
 LM & \textbf{65.3} & \textbf{28.1} & \textbf{9.70} & 6.30 & 3.60 & 1.70 & 0.25 \\
 PMI & 1.80 & 3.30 & 3.36 & 4.10 & 4.00 & 4.90 & 7.00 \\
 \hline
\end{tabular}
\caption{Recall@100 Narrative Cloze Results. $<C$ indicates that instances whose cloze answer is one of the top $C$ most frequent events are not evaluated on}
\label{table:cloze_recall50}
\end{table*}

The narrative cloze task, or some variant of it, has remained a popular automatic test for systems aiming to extract `script' knowledge. The task is usually formulated as follows: given a chain of events $e_1, ... e_{n-1}$ that occurs in the data, predict the held out next event that occurs in the data, $e_n$. There exists various measures to calculate a models ability to perform in this task, but arguably the most used one is the Recall@N measure introduced in \citet{jans2012skip}. Recall@N works as follows: for a cloze instance, a system will return the top N guesses for $e_n$. Recall@N is the percentage of times $e_n$ is found anywhere in the top N list.

The automatic version of the cloze task has notable limitations. As noted in \citet{Rudinger2015}, the cloze task is essentially a language modeling task; it measures how well a model \textit{fits} the data. The question then becomes whether data fit implies valid script knowledge was learned. The work of \citet{chambers2017behind} casts serious doubts on this, with various experiments showing automatic cloze evaluations are biased to high frequency, uninformative events, as opposed to informative, \textit{core}, script events. They further posit human annotation as a necessary requirement for evaluation. 

In this experiment, we provide another datapoint for the inadequacy of the automatic cloze, while simultaneously showing the relative robustness of the knowledge extracted from our Causal system. For the experiment, we make the following assumptions: (1) Highly frequent events tend to appear in many scenarios, and hence are less likely to be a informative `core' event for a script (such as `pay' or  `shoot'), and (2) Less frequent events are more likely to appear only in specific scenarios, and are thus more likely to be informative events. If these are true, then a system that has extracted useful script knowledge should keep (or even improve) performance on the cloze when the correct answer for $e_n$ is a less frequent event. 

We thus propose a Infrequent Cloze task. In this task we create a variety of different cloze datasets (each with 2000 instances) from our test set. Each set is indexed by a value $C$, such that the indicated dataset does not include instances from the top $C$ most frequent events ($C=0$ is the normal cloze setting). We compute a Recall@100 cloze task on 7 sets of various $C$ and report results in Table \ref{table:cloze_recall50}.

At $C=0$, as expected, the LM model is vastly superior. The performance of the LM model drastically drops however, as soon as $C$ increases, indicating an overreliance on prior probability. The LM performance drops below 2\% once $C=200$, indicating almost no ability in predicting informative events such as \textit{drink} or \textit{pay}, both of which occur in this set in our case. 

The PMI and Causal model's performance on the other hand, steadily improve while $C$ increases, with the Causal model consistently outperforming PMI. This result, \emph{when combined with} the results of the human evaluation, give further evidence towards the relative robustness of the Causal model in extracting informative core events. The precipitous drop in performance of the LM further underscores problems that a naive automatic cloze evaluation may cover up.

\section{Related Work}
Our work looks at script like associations between events in a manner similar to \citet{chambers:08}, and works along similar lines \cite{jans2012skip,Pichotta2016}. Related lines of work exist, such as work using generative models to induce probabilistic schemas\cite{chambers2013event,cheung2013probabilistic,ferraro2016unified}, and other work showing how script knowledge may be mined from user elicited event sequences \cite{regneri2010learning, orr2014learning}. The cognitive linguistics literature is rich with work studying the role of causal semantics in linguistic constructions and argument structure~\cite{talmy1988force, croft1991syntactic,croft2012verbs}, as well as the causal semantics of lexical items themselves \cite{wolff2003models, Wolff2007}. Work in the NLP literature on extracting causal relations has benefited from this line of work, utilizing the systematic way in which causation in expressed in language to mine relations \cite{girju2002mining,Girju2003,Blanco2008,Do2011,Bosselut2019}. This line work aims to extract causal relations between events that are in some way explicitly expressed in the text (e.g. through the use of particular constructions).
Taking advantage of how causation is expressed in language may benefit our causal model, and is a potential path for future work.


\section{Conclusions and Future Work}
In this work we argued for a causal basis in script learning. We showed how this causal definition could be formalized and used in practice utilizing the tools of causal inference, and verified our method with human and automatic evaluations. In the current work, we showed a method calculating the `goodness' of a script in the simplest case: between pairwise events, which we showed still to be quite useful. A causal definition is in no way limited to this pairwise case, and future work may generalize it to the sequential case or to event representations that are compositional (for example, by performing multiple interventions). Having a causal model shines a light on the assumptions made here, and indeed, future work may further refine or overhaul them, a process which may further shine a light on the nature of the knowledge we are after.

\bibliography{causal_events}
\bibliographystyle{acl_natbib}
\section{Appendix}
%

\subsection{Data Pre-Processing}
For these experiments, we use the Toronto Books corpus \cite{zhu2015aligning,NIPS2015_5950}, a collection of fiction novels spanning multiple genres.
The original corpus contains 11,040 books by unpublished authors.
We remove duplicate books from the corpus (by exact file match), leaving a total of 7,101 books; a distribution by genre is provided in Table \ref{tab:tbks}.
The books are assigned randomly to train, development, and test splits in 90\%-5\%-5\% proportions (6,405 books in train, and 348 in development and test splits each).
Each book is then sentence-split and tokenized with CoreNLP 3.8 \cite{manning-EtAl:2014:P14-5}; these sentence and token boundaries are observed in all downstream processing.

\begin{table}
\centering
\begin{tabular}{|lr|lr|}
\hline
Adventure 	&   390 & Other				&   284 \\
Fantasy		& 1,440 & Romance			& 1,437 \\
Historical	&   161 & Science Fiction	&   425 \\
Horror		&   347 & Teen				&   281 \\
Humor		&   237 & Themes			&    32 \\
Literature	&   289 & Thriller			&   316 \\
Mystery		&   512 & Vampires			&   131 \\
New Adult	&   702 & Young Adult		&   117 \\
\hline
\end{tabular}
\caption{Distribution of books within each genre of the deduplicated Toronto Books corpus.}
\label{tab:tbks}
\end{table}

\subsubsection{Narrative Chain Extraction Pipeline}
In order to extract the narrative chains from the Toronto Books data, we implement the following pipeline.
First, we note that coreference resolution systems are trained on documents much smaller than full novels \cite{pradhan2012conll}; to accommodate this limitation, we partition each novel into non-overlapping windows that are 100 sentences in length, yielding approximately 400,000 windows in total.
We then run CoreNLP's universal dependency parser \cite{NIVRE16.348,chen-manning-2014-fast}, part of speech tagger \cite{toutanova2003feature}, and neural coreference resolution system \cite{clark-manning-2016-deep,clark-manning-2016-improving} over each window of text.
For each window, we select the longest coreference chain and call the entity in that chain the ``protagonist,'' following \citet{chambers:08}.

We feed the resulting universal dependency (UD) parses into PredPatt \cite{uds2016}, a rule-based predicate-argument extraction system that runs over universal dependency parses.
From PredPatt output, we extract predicate-argument edges, i.e., a pair of token indices in a given sentence where the first index is the head of a predicate, and the second index is the head of an argument to that predicate.
Edges with non-verbal predicates are discarded.

At this stage in the pipeline, we merge information from the coreference chain and predicate-argument edges to determine which events the protagonist is participating in.
For each predicate-argument edge in every sentence, we discard it if the argument index does not match the head of a protagonist mention.
Each of the remaining predicate-argument edges therefore represents an event that the protagonist participated in.

With a list of PredPatt-determined predicate-argument edges (and their corresponding sentences), we are now able to extract the narrative event representations, $(p,d)$
For $p$, we take the lemma of the (verbal) predicate head.
For $d$, we take the dependency relation type (e.g., \textit{nsubj}) between the predicate head and argument head indices (as determined by the UD parse); if a direct arc relation does not exist, we instead take the unidirectional dependency path from predicate to argument; if a unidirectional path does not exist, we use a generic ``arg'' relation.

To extract a factuality feature for each narrative event (i.e. whether the event happened or not, according to the meaning of the text), we use the neural model of \citet{rudinger-etal-2018-neural}.
As input to this model, we provide the full sentence in which the event appears, as well as the index of the event predicate's head token.
The model returns a factuality score on a $[-3,3]$ scale, which is then discretized using the following intervals: $[1,3]$ is ``positive'' ($+$), $(-1,1)$ is ``uncertain,'' and $[-3,-1]$ is ``negative'' ($-$).


From this extraction pipeline, we yield one sequence of narrative events (i.e., narrative chain) per text window.

\subsection{Training and Model Details - Causal Model}
\subsubsection{RNN Encoder}
We use a single layer GRU based RNN encoder with a 300 dimensional hidden state and 300 dimensional input event embeddings to encode the previous events into a single 300 dimensional vector. 
\subsubsection{CNN Encoder} 
We use a CNN to encode the text into a 300 dimensional output vector. The CNN uses 4 filters with ngram windows of $(2,3,4,5)$ and max pooling.
\subsubsection{Training Details - Pretraining}
The conditional for the Causal model is trained using Adam with a learning rate of 0.001, gradient clipping at 10, and a batch size of 512. The model is trained to minimize cross entropy loss. We train the model until loss on the validation set does not go down after three epochs, afterwhich we keep the model with the best validation performance, which in our case was epoch 4
\subsubsection{Training Details - Finetuning}
The model is then finetuned on our dataset of 2000 annotated examples. We use the same objective as above, training using Adam with a learning rate of 0.00001, gradient clipping at 10, and a batch size of 512. We split our 2000 samples into a train set of 1800 examples and a dev set of 200 examples. We train the model in a way similar to above, keeping the best validation model (at epoch 28).

\subsection{Training and Model Details - LM Baseline}
We use a 2 layer GRU based RNN encoder with a 512 dimensional hidden state and 300 dimensional input event embeddings as our baseline event sequence LM model.
\subsubsection{Training Details} 
The LM model is trained using Adam with a learning rate of 0.001, gradient clipping at 10, and a batch size of 64. We found using dropout at the embedding layer and the output layers to be helpful (with dropout probability of 0.1). The model is trained to minimize cross entropy loss. We train the model until loss on the validation set does not go down after three epochs, afterwhich we keep the model with the best validation performance, which in our case was epoch 5.

\subsection{Annotation Interfaces}
\begin{figure}[t!]
\centering
\includegraphics[scale=0.26]{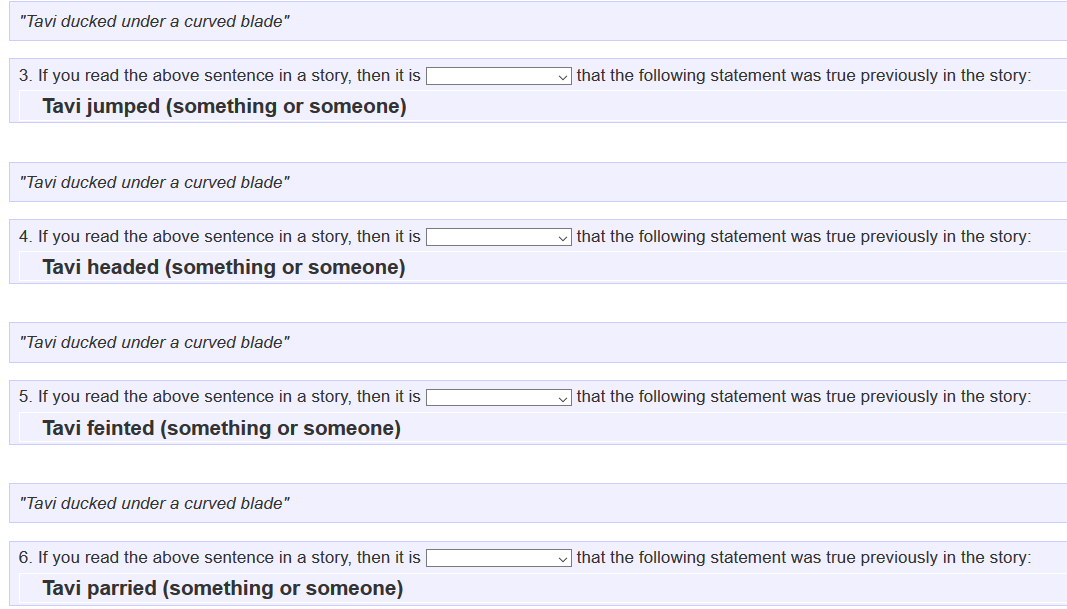}
\caption{\label{fig:candsinterface}The annotation interface for the out-of-text events annotation.} 
\end{figure}

\begin{figure}[t!]
\centering
\includegraphics[scale=0.32]{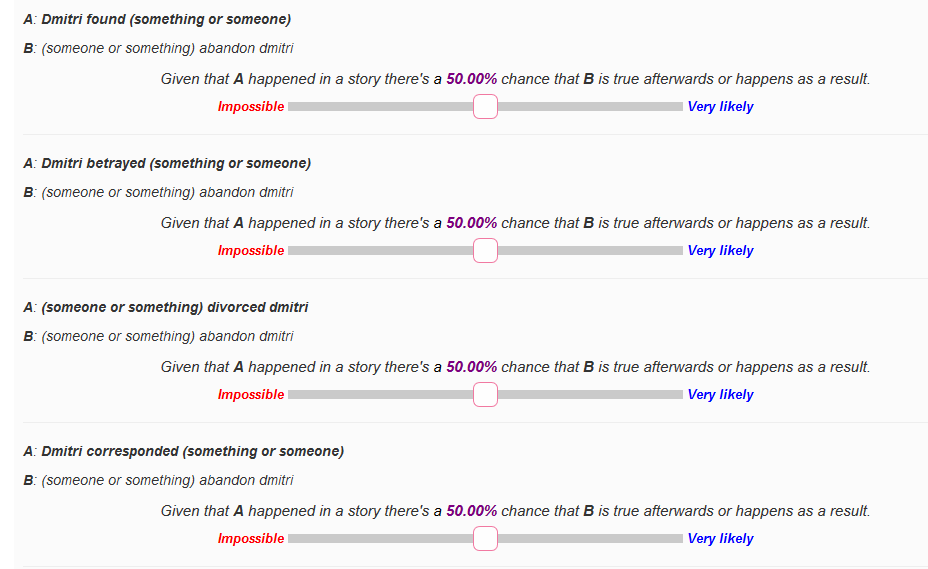}
\caption{\label{fig:PairwiseInterface}The annotation interface for the pairwise human evaluation annotation experiment.} 
\end{figure}

\begin{figure}[t!]
\centering
\includegraphics[scale=0.32]{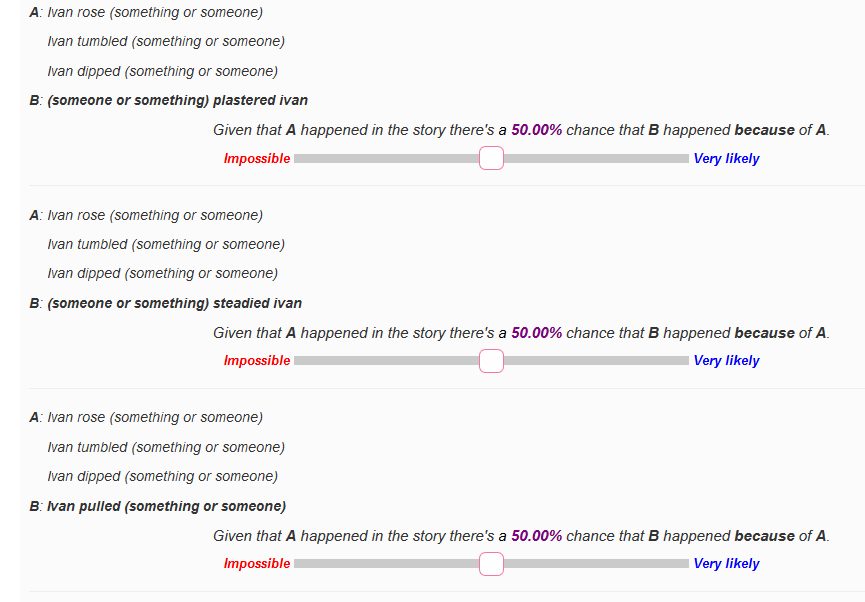}
\caption{\label{fig:chaininterface}The annotation interface for the chain completion human evaluation annotation experiment.} 
\end{figure}

To get an idea for about the annotation set ups used here, we also provide screen shots of the annotation suites for all three annotation experiments. The out-of-text annotation experiment of Section 3.3 is shown in Figure \ref{fig:candsinterface}. The pairwise annotation evaluation of Section 4.3 is shown in Figure \ref{fig:PairwiseInterface}. The chain completion annotation evaluation of Section 4.4 is shown in Figure \ref{fig:chaininterface}.

\end{document}